\documentclass[runningheads]{llncs}
\usepackage{xcolor}
\usepackage{hyperref}
\usepackage{cite}
\usepackage{array}
\usepackage{multirow}
\usepackage[normalem]{ulem}
\usepackage[T1]{fontenc}
\usepackage{graphicx}
\usepackage{subcaption}
\usepackage{amsmath}

\begin{document}
\title{Matching Problems to Solutions: An Explainable Way of Solving Machine Learning Problems}
\titlerunning{Explainable ML Problem Solving}
\author{Lokman Saleh\inst{1}\orcidID{0000-1111-2222-3333} \and
Hafedh Mili\inst{1}\orcidID{2222--3333-4444-5555}\and 
Mounir Boukadoum\inst{1}\orcidID{1111-2222-3333-4444} \and
Abderrahmane Leshob\inst{1}\orcidID{1111-2222-3333-4444}}
%
%\authorrunning{L. S. et al.}
\institute{
 LATECE Lab, Universit\'e du Qu\'ebec \`a Montr\'eal, H3C 3P8, Montr\'eal, Canada \\
  \url{http://www.latece.uqam.ca}\\
  \email{saleh.lokman@courrier.uqam.ca, boukadoum.mounir@uqam.ca, mili.hafedh@uqam.ca} 
}
\maketitle              % typeset the header of the contribution
\begin{abstract}
Domain experts from all fields are called upon, working with data scientists, to explore the use of ML techniques to solve their problems. Starting from a domain problem/question, ML-based problem-solving typically involves three steps:  (1) formulating the business problem (\textit{problem domain}) as a data analysis problem (\textit{solution domain}), (2) sketching a high-level ML-based solution pattern, given the domain requirements and the properties of the available data, and (3) designing and refining the different components of the solution pattern. There \textit{has to be} a \textit{substantial body of ML problem solving knowledge} that ML researchers agree on, and that ML practitioners routinely apply to solve \textit{the most common problems}. Our work deals with capturing this body of knowledge, and embodying it in a ML problem solving workbench to helps domain specialists who are not ML experts to explore the ML solution space. This paper focuses on: 1) the \textit{representation} of domain problems, ML problems, and the main ML solution artefacts, and 2) a heuristic matching function that helps identify the ML \textit{algorithm family} that is most appropriate for the domain problem at hand, given the domain (expert) requirements, and the characteristics of the training data. We review related work and outline our strategy for validating the workbench.

\keywords{Domain Engineering \and Machine learning \and ML Workbench.}
\end{abstract}
\section{Introduction}
\label{sec:introduction}
Recent advances in ML algorithms and computer hardware, and the increasing availability of domain data have led to the proliferation of ML solutions in many application domains, including  health care, marketing, autonomous vehicles, precision agriculture, smart infrastructures, and others. A medical doctor who treats Alzheimer and has at her/his disposal a large database of Alzheimer patient records that include the results of biological tests taken several years prior to the onset of Alzheimer, may be interested in identifying precursor markers. A major software vendor that has a niche product in its offering, uses machine learning to target a marketing campaign towards the kinds of companies that have historically bought its product. Typically, domain experts working with data scientists are called upon to explore the potential of ML techniques to solve a problem specific to their domain. A data scientist would typically perform the following steps in sequence: (1) translate the business problem or question into a data analysis problem, (2) sketch the general workflow of a ML-based solution, given the needs of the domain expert and the properties of the available data, and (3) design and refine the different components of the workflow based on the previous two steps, and on experimentation. Our work is concerned with the development of tools to assist with the three tasks.

There exists a \textit{large body of ML problem-solving knowledge} that ML researchers agree on, and that ML practitioners routinely apply to solve \textit{the most routine problems}, regarding at least the last two steps: (1) identifying the general outline of an \textit{ML-based solution pattern}, given a \textit{data analysis problem}, domain requirements, and data characteristics; and (2) the \textit{selection} and \textit{specialization} of the various components of the solution pattern. Our goal is to capture this knowledge, and embody it in an ML problem-solving workbench that helps domain specialists who are not ML experts to explore the ML solution space. 

This paper outlines the general approach, presented in Section \ref{sec:towards-ML-based-solution-workbench}, but is concerned primarily with the identification and codification of the ML problem solving knowledge (Section \ref{sec:model-ML-problems-solutions}), and identifying the general outline of an \textit{ML-based solution pattern} (Section \ref{sec:matching-problems-solutions}). A key aspect of any ML-based solution pattern is the identification of the \textit{algorithm family} that is most appropriate the ML-problem at hand (e.g. classification vs. regression), the domain (expert) prioritized requirements (e.g. explainability vs. accuracy), and the characteristics of the data (size, quality, statistical properties, etc.). In particular, we present a \textit{heuristic matching function} that helps rank candidate algorithm families by degree of fitness for the problem at hand, by presenting the relevant characteristics of algorithm families (Section \ref{subsec:matching-problems-solutions:selection-criteria-algorithm}), problem requirements (Section \ref{subsec:matching-problems-solutions:ML-problem-requirements}), the correspondences between the two (Section \ref{subsec:matching-problems-solutions:mapping-requirements-selection-criteria}), the desirable mathematical properties of the matching function (Section \ref{subsec:matching-problems-solutions:desirable-properties-matching-function}), and then the matching function (Section \ref{subsec:matching-problems-solutions:matching-function}.

We present our validation strategy in Section \ref{sec:validation-strategy}, related work in Section \ref{sec:related-work}, and conclude in Section \ref{sec:discussion}.

\section{Towards a ML-based Problem Solving Workbench}
\label{sec:towards-ML-based-solution-workbench} 
\begin{figure} 
        \centering
        \includegraphics[width=\textwidth]{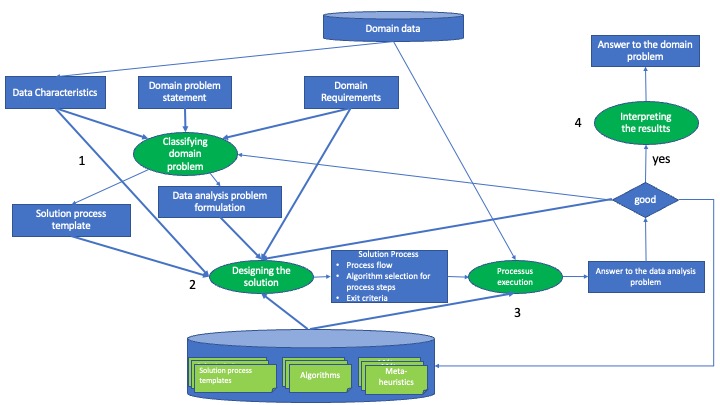}
        \caption{Workflow for ML-based problem solving}
        \label{fig:ML-problem-solving-workflow}
\end{figure}

Domain experts are called upon, alone or working with data scientists, to explore the potential of ML techniques to solve a domain problem, often framed as a question. In medical applications, the question may be about the presence or absence of a specific pathology, e.g. from an X-Ray. For a bank, it can be whether a customer is likely to default on a loan (a binary classification) based on current or potential credit history; another question might be which aspects of the credit history are the best predictors of default (principal component analysis). A retailer might ask 'who buys product X?' (n-ary classification); yet another might be 'can I tell the projected lifetime value of a customer from their purchase history?' (a regression), etc.  Data scientists would first reframe the business problem in terms of a \textit{data analysis problem}--put between parentheses in the previous examples. Then, they draw on their expertise and what they learn from the domain expert about the structure and quality and the data, and the domain-specific requirements, to \textit{first} sketch a solution pattern, which they can later refine based on additional information about the domain and the data. This is step 1 in Figure \ref{fig:ML-problem-solving-workflow}.

\begin{figure}
        \centering
        \includegraphics[width=\textwidth]{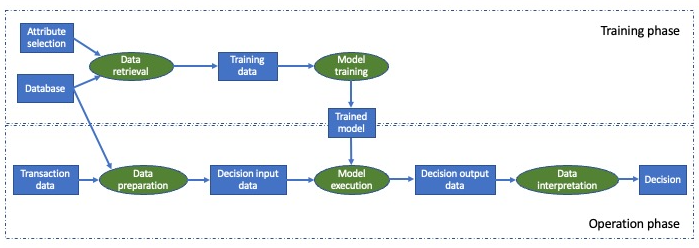}
        \caption{A sample generic solution template}
        \label{fig:sample-generic-solution-template}
\end{figure}

Knowing that a problem is about classification with labeled testing data may identify a \textit{general class} of machine learning algorithms (e.g. a \textit{supervised classification algorithm}), and a generic problem solving workflow, but more needs to be done  \ref{fig:ML-problem-solving-workflow} shows one such workflow. The data scientist needs to refine this workflow by, a) identifying the required substeps, and selecting the appropriate algorithm/procedure for each substep. This is step 2 in Figure \ref{fig:ML-problem-solving-workflow}. Activities in this step may include, 1) eliciting the domain requirements (e.g. from the domain expert), and 2) learning more about the data. For example, in regulated domains such as financial services, auditability of decisions is important, which excludes black box approaches. Similarly, for medical diagnosis based on X-ray imagery, accuracy is most valued, whereas for a customer experience application (e.g. selecting ads for a landing page), speed is most important. Regarding the training data, we may want to know about its structure, type (numerical vs categorical), size, quality, sparsity, presence of noise, missing data, and others, as different algorithm families are known to better suited for different data characteristics. Using this information, the data analyst can refine the workflow illustrated in Figure \ref{fig:sample-generic-solution-template} by breaking down the generic steps into detailed substeps. For example, they may include input data retrieval, filtering, transformation (from categorical to numerical and vice-versa), missing value handling, denoising, normalization, dimensionality reduction, and so forth, as deemed appropriate for the domain requirements, the data characteristics and the prerequisites of the model-building algorithm family. The interplay of these various criteria reduces the algorithmic possibilities for each processing step.

After a first detailed solution, we start an iterative phase to build a model for the solution process--corresponding to the upper part of the generic process flow of Figure \ref{fig:sample-generic-solution-template}--until meeting the \textit{exit criteria}, for example getting a performance metric (precision, recall, speed, etc.) above a threshold set by the domain expert. This corresponds to step 3 in Figure \ref{fig:ML-problem-solving-workflow}, with a feedback loop to revise the detailed design (step 2) or the identification of the general algorithm family (step 1). 

After achieving the desired performance, we tackle the interpretation of the results to make them digestible for domain users, which is needed for deploying the solution. This corresponds to refining/finalizing the last step of the \textit{operation process}, corresponding to the lower part of Figure \ref{fig:sample-generic-solution-template}. This is represented by step 4 in Figure \ref{fig:ML-problem-solving-workflow}. If the iterative solution refinement (step 3 of Figure \ref{fig:ML-problem-solving-workflow}) does not get us  near or above the minimal performance requirements, we may give up on a machine learning solution, and skip step 4 altogether. 

Ultimately, we aim to provide tool assistance for all the four previous steps. At this stage of our work, we are focusing on the first two steps, 1) the classification of the business problem into a data analysis problem, and 2) the initial design of the solution. This paper focuses on step 2.
\section{A model of ML problems and solutions}
\label{sec:model-ML-problems-solutions}
\subsection{Concept of operations}
\label{subsec:model-ML-problems-solutions:concept-of-operations}
\begin{figure}
        \centering
        \includegraphics[width=0.67\textwidth]{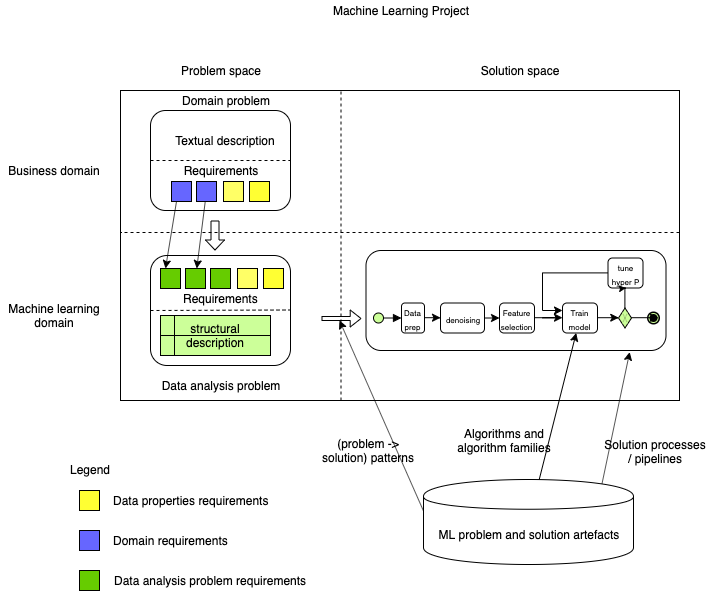}
        \caption{The concept of \textit{Machine Learning Project} in our ML problem solving workbench}
        \label{fig:concept-of-operations}
\end{figure}

Our \textit{Machine Learning (ML) Problem Solving Workbench} needs to support the ML problem solving process presented in the previous section (see Figure \ref{fig:ML-problem-solving-workflow}). It does so by manipulating and managing a \textit{Machine Learning (ML) project} through the four steps/stages outlined above. As illustrated in Figure \ref{fig:ML-problem-solving-workflow}, this should include both the initial formulation of the \textit{domain problem}, and the machine learning (ML solution), as it evolves through steps (2) and (3). This is represented graphically in Figure \ref{fig:concept-of-operations}.

An ML project is born with the formulation of the domain problem/question which, for the time being, consists of: 1) a textual description of the business problem, and 2) a bunch of requirements presented as key-value pairs. For the time being, we assume that the textual description is free-form, with no explicit structure; in future versions, as we tackle the mapping from domain problems to data analysis problems, we may think of a more structured input relying on problem formulation aids. The requirements are of two types: a) \textit{domain requirements} (blue boxes), including things such accuracy, audibility, speed, transparency, etc., and b) \textit{data requirements} (yellow boxes) which embody \textit{data properties} that have a significant influence on the algorithm family, including whether the data is labeled or not, size, representativeness, noisiness, etc. Some of these properties will be provided by the domain expert (e.g., the labeling, representativeness) while others can be assessed by our workbench, given access to the data \footnote{In the current implementation, the user has the option of provide the URL and credentials of a relational data source, and the tool will assess a number of properties, including size, missing values, statistical distributions, etc.}.
This representation of the \textit{domain problem} is then \textit{somehow} mapped to a \textit{data analysis problem} (lower left corner of \ref{fig:concept-of-operations}); how the mapping works is beyond the scope of this paper. This shows that the data properties are carried over unchanged, whereas the \textit{domain requirements} are mapped to corresponding \textit{data analysis requirements} (green boxes). For example, the \textit{business requirement} of \textit{auditability} in financial services industries, is translated into a data analysis/ML \textit{explainability} requirement. The \textit{structural description} will be discussed in the next section.

As mentioned in Section \ref{sec:towards-ML-based-solution-workbench}, the formulation of the \textit{data analysis problem} (e.g., "supervised multi-class classification, with a small data set"), will identify an initial solution pattern/sketch, which can then be refined by an iterative process where the components of the project influence each other in many way:
\begin{enumerate}
    \item The identification of an algorithm family may dictate a number of preprocessing and post-processing steps.
    \item The specification of additional data properties, or modification of existing ones. For example, simple statistical properties are evaluated systematically (e.g. missing values, average, standard deviation) whereas more costly ones are evaluated on-demand. 
    \item The modification of existing properties (values or importance). For example, a domain user may \textit{first} require explainability and high accuracy, with equal weight, and then realizing that no algorithm family scores highly for both, they may change the relative importance of explainability.
\end{enumerate}
To this end, the workbench relies on a \textit{repository of machine learning problem and solutions artefacts}, seen at the bottom of Figure \ref{fig:concept-of-operations}. We will say a bit more about the structure of this repository in the next two sections. The curious reader can check a work-in-progress version of our workbench at \url{https://isolvemymlproblem-c6d96d0c8560.herokuapp.com/login}.
\subsection{The ML Project Metamodel}
\label{subsec:model-ML-problems-solutions:ML-project-metamodel}
\begin{figure}
        \centering
        \includegraphics[width=0.6\textwidth]{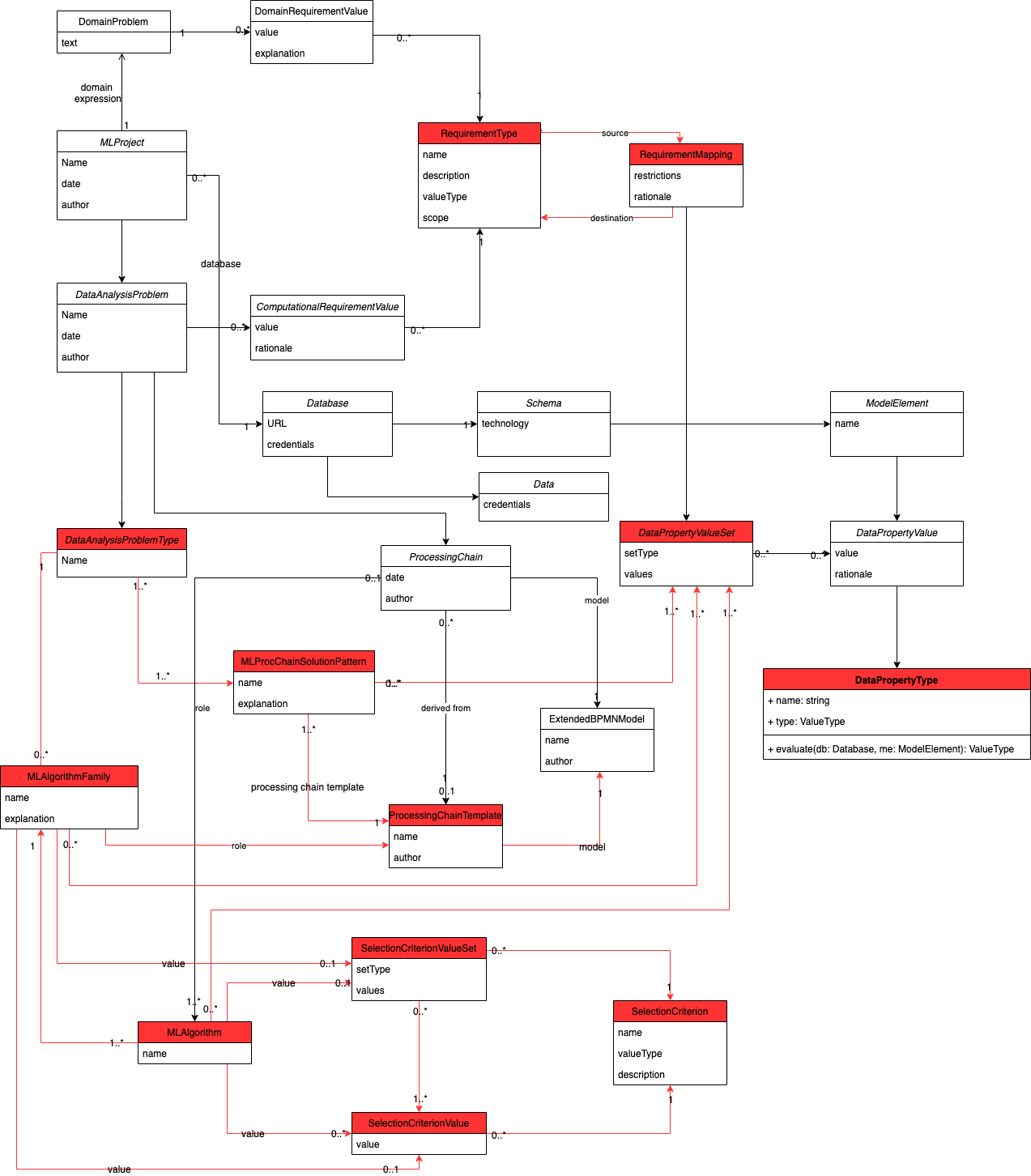}
        \caption{A UML model for machine learning problems and solutions}
        \label{fig:ML-project-metamodel}
\end{figure}

Figure \ref{fig:ML-project-metamodel} shows the UML ML project model. We will highlight the main elements of the model that relate to Section \ref{subsec:model-ML-problems-solutions:concept-of-operations} (Figure \ref{fig:concept-of-operations}). To help understand the metamodel, we colored in \textit{red} the entities and associations that are part of the repository of ML problem and solution artefacts, to distinguish them from 'instance data', i.e. constructs specific to a ML project \textit{instance}.

Starting from the top-left corner, an \textbf{MLProject} includes: 1) a domain problem formulation, represented by the class \textbf{DomainProblem}, and the one-to-many association to \textbf{DomainRequirementValue} (property value), which is linked to its \textbf{RequirementType}, and 2) a data analysis problem formulation, represented by the 'head class' \textbf{DataAnalysisProblem}, also associated with a bunch of requirements represented by \textbf{ComputationalRequirementValue}s, which are also mapped to \textbf{RequirementType}. The \textbf{RequirementMapping} class is used to link \textit{domain requirements} (e.g. auditability) to \textit{computational requirements} (e.g. explainability).
The representation of the database associated with the project is shown further down in Figure \ref{fig:ML-project-metamodel}. In the current implementation, we rely on the availability and accessibility of a relational database (URL and credentials). The tool connects the user to the base, retrieves the schema, and prompts the user to select the attributes to include on the training dataset. For the purposes of this paper, we are not detailing the representation of the schema, but suffice it to say that a schema includes a set of \textbf{DataElement}s, which correspond to the data attributes (table columns) that the user selects for the model training. For each such \textbf{DataElement}, we will have data properties, represented by \textbf{DataPropertyValue} $\rightarrow$ \textbf{DataPropertyType} pairs. For example, a \textbf{DataPropertyType} may be \textit{Has Missing Values}, with possible values (instances of \textbf{DataPropertyValue}) \textit{True} and \textit{False}. Statistical properties may be represented in a similar fashion.

The representation of algorithms (families) and processing chains/pipelines is discussed next.
\subsection{Representing algorithms and processing pipelines}
\label{subsec:model-ML-problems-solutions:representing-algorithms-processing-pipelines}
As mentioned in Section \ref{sec:towards-ML-based-solution-workbench}, ML \textit{model building} is \textit{one}--albeit central--step in a sequence. This was illustrated in Figure \ref{fig:sample-generic-solution-template}, where the process task "Data Retrieval" involves several steps. Further, the "Model Training" task in that same Figure involves a non trivial process. We called the sequence of steps, starting with data ingestion to model 'delivery' a \textit{processing chain/pipeline}. Thus, we start by describing processing chains (next), then talk about the representation of algorithms.
\subsubsection{Representing algorithms and algorithm families - principles}
\label{subsubsec:model-ML-problems-solutions:representing-algorithms-processing-pipelines:algorithms}
Our representation of ML algorithms and families needs to support the users of our benchmark in steps (2), (3), and (4) explained in Section \ref{sec:towards-ML-based-solution-workbench}. Steps (3) and (4) involve \textit{executing} these algorithms, which assumes that we have access to executable versions thereof. This is not an issue as there are many machine learning libraries in different languages (C\#, Java, Python, etc.), each reflecting the idiosyncrasies of the implementation language and the API design choices of its authors. Thus, we chose \textit{not} to represent the kind of information needed to execute the algorithms, because: a) that documentation depends on the library at hand, \textit{and} b) it is available in such libraries themselves. Accordingly, we focused on representing algorithms at a level that supports \textit{solution design}, i.e. designing the processing chain.

Selecting the algorithm family, itself, depends heavily on: 1) the type of data analysis problem (e.g. regression versus clustering versus binary classification), and 2) a handful of domain requirements and data properties. This is the reason behind \textit{cheat sheets} such as \cite{Microsoft} or \cite{SAS2020}, enabling us to quickly narrow down the set of algorithms than can be used for any step of the processing chain.
Accordingly, we decided to represent algorithms and algorithm families with no structural or behavioral information, but with a flat set of properties, referred to in the ML literature as \textit{selection criteria} or \textit{selection drivers}. This is embodied in the classes \textbf{SelectionCriterion} (e.g. \textit{explainability}), \textbf{SelectionCriterionValue} (e.g. \textit{true} or \textit{false}), and \textbf{SelectionCriterionValueSet}. The latter class is needed in case a single algorithm has several values for the same selection criterion, or to account for the fact that the \textit{members of an algorithm family} have different values for the criterion.

\subsubsection{Representing processing chains/pipelines - principles}
\label{subsubsec:model-ML-problems-solutions:representing-algorithms-processing-pipelines:processing-chains}
\begin{figure}
        \centering
        \includegraphics[width=0.8\textwidth]{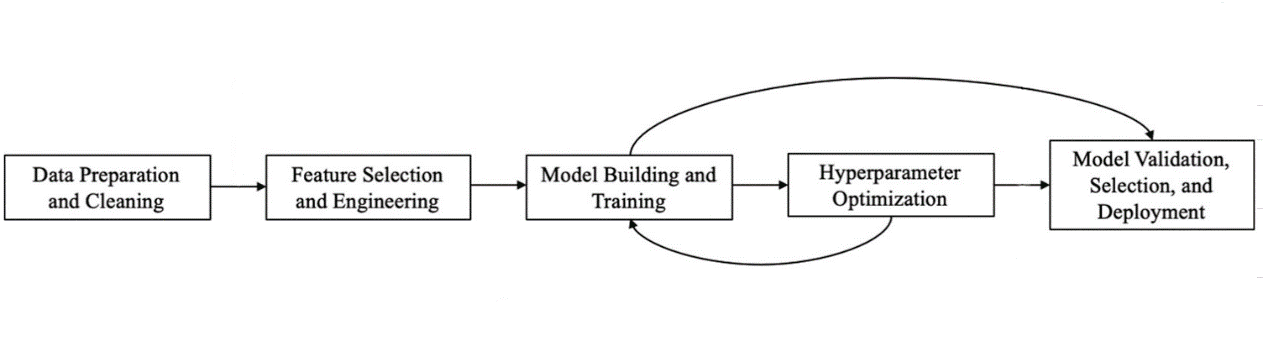}
        \caption{A sample model building processing chain.}
        \label{fig:sample-processing-pipeline}
\end{figure}

Figure \ref{fig:sample-processing-pipeline} shows a sample processing chain, represented at a fairly coarse and informal level. Different authors have proposed different model building \textit{processes}, but they typically include sequences of steps with branching depending on some tests, iteration, and other common control constructs. Differences between the published processes are due to a variety of factors, including: 
\begin{enumerate}
    \item What each author considers as the starting and end points of the process. For example, in our case, data(base) access is part of the problem definition, whereas some authors might included a data ingestion step.
    \item The granularity of the represented tasks. Figure \ref{fig:sample-processing-pipeline} shows a single step for 'Data Preparation and Cleaning', whereas a different author might separate the two steps.
    \item The data used for model training, where certain types might require preprocessing--e.g. word embedding for textual data.
    \item The actual model building algorithm, which may only take data in a specific format (e.g. categorical versus numerical), with potential pre and post-processing steps depending on the data and the application domain.
\end{enumerate}

We mentioned in Section \ref{sec:towards-ML-based-solution-workbench} that the outcome of step (2) (detailed design) is a processing chain where each step is bound to a specific algorithm. To evaluate such a solution, the chain must be \textit{executed}.

All in all, we need a visual (domain users), executable action language, that supports nesting and different levels of abstraction. BPMN satisfies all of these requirements. We also needed to add some custom behavior to the \textit{tasks}, including:
\begin{enumerate}
    \item Showing the set of algorithms/library functions that can play the role indicated by each task, and
    \item Dynamically updating that set when computational requirements are modified, or new data properties are computed
    \item Supporting other machine-learning specific tasks.  
\end{enumerate}
Accordingly, we decided to represent our processing chains with an extension of BPMN2 that specializes BPMN2's \textit{service tasks}.

\section{Matching problems to solutions}
\label{sec:matching-problems-solutions}
\subsection{Selection criteria for algorithms and algorithm families}
\label{subsec:matching-problems-solutions:selection-criteria-algorithm}
To determine the list of criteria to use, we started with a review of the literature on ML algorithms, and more specifically, those papers that investigated the appropriate families of algorithms to solve a specific class of ML or domain problems, which identified algorithm family \textit{selection criteria} \cite{48andreopoulos2009roadmap, 57das2017survey, fatima2017, patil2014sentiment}. From there, we went through the following iterative process:
\begin{enumerate}
    \item Take the simple union of the selection criteria proposed in the literature. We identified no fewer than 40 criteria.
    \item Make a first pass through the list to identify duplicates under different names, and remove them. In our case, this reduced the list by five.
    \item Make a second pass to reduce the list to one of \textit{orthogonal properties}. Indeed, among the \textit{unique} selection criteria (output of step 2), we encountered cases of selection criteria that had dependencies, e.g. when one criterion entails the other or when one is the exact opposite of the other (e.g. \textit{resilience to noise} vs. \textit{sensitivity to noise}). This removed two selection criteria--down to 33.
    \item We went through the list to identify or keep only the most salient features, in part to facilitate knowledge entry. We ended up with 27.   
\end{enumerate}
Table \ref{tab:criteres-selection-poids} shows the list of criteria.

For each selection criterion, we included a weight and a value range. 
Weights can be seen as a measure of the effort it takes to compensate for failing the requirement. Looking at properties with weight 'A', there is no remedy for not having the appropriate training type (\textit{Training type}). Idem for \textit{Explainability}: that is an \textit{intrinsic} property of the \textit{structure} of the ML model; a decision tree is \textit{inherently} explainable whereas a deep neural network is not.

As we go down the weight scale (B, C, and D), if an algorithm does not satisfy the corresponding property 'out of the box', we can remedy the situation by adding a more (B) or less (D) onerous preprocessing step. Take the example of \textit{Tolerance to correlated attributes}. If an algorithm family is known to be derailed by correlated attributes, and we know for fact, or fear that the input data includes correlated attributes, then we can remedy the situation by adding a \textit{principal component analysis} preprocessing step. Going down the list to \textit{Tolerance to noise} (C), if an algorithm family has low, or no, tolerance to noise, than we can simply 'denoise' the data before submitting it for model training. 

There are some properties that cannot be 'fixed', such as \textit{Incrementality}, with a weight of C. This is an intrinsic property of the ML model, or its building algorithm, that cannot be 'fixed' with preprocessing. At worst, it is a 'C level annoyance', but we can always retrain our model fully with the old and new data, at tolerable time intervals.

Both weights and value ranges are given as \textit{linguistic} (\textit{fuzzy}) values, with the exception of \texttt{Accuracy}. Depending on the final form of the matching function, they can later be converted to numbers.

\begin{table}
    \centering
    \begin{tabular}
    {| m{6cm} | c | m{5cm} |}
        \hline
        \textbf{Selection criterion} & \textbf{Weight}  & \textbf{Value range}\\

    \hline  
    \hline
        Training type  & A & \{supervised, unsupervised, reinforcement,...\} \\ 
    \hline
        Explainability (how the algorithm arrives at the result) & A & \{Explainable, Not explainable \}\\
    \hline
        Interpretability (what the result means) & A-B & \{Interpretable, Not interpretable \}\\
    \hline
        Model building's sensitivity to input order & A-B & \{None, Low, Medium, High \}\\
    \hline
        Accuracy & B & \{$<= ~ 80\%$, $[$80\%, 90\%$]$, $>= ~ 90\%$ \}\\
    \hline
        Tolerance to correlated attributes & B & \{None, Low, Medium, High \} \\
    \hline
        Resilience to overfitting & B & \{None, Low, Medium, High \}\\
    \hline
        Tolerance to data imbalance & B & \{None, Low, Medium, High \}\\
    \hline
        Ease of hyper-parameter setting & B & \{Low, Medium, High \}\\
    \hline
        Model training complexity & B & \{Low, Medium, High \}\\
    \hline
        Ability to handle multiple classes & B & \{Yes, No \}\\
    \hline
        Volume of data required for convergence & B & \{Low, Medium, High \}\\
    \hline
        Ability to handled highly dimensional data & B-C & \{Yes, No \}\\
    \hline
        Ability to handle missing records or attributes & B-C& \{None, Low, Medium, High \} \\
    \hline
        Incrementality (ability to integrate new data incrementally) & C & \{Yes, No \}\\
    \hline
        Transparency & C & \{Yes, No \}\\
    \hline
        Tolerance to noise & C & \{Low, Medium, High \}\\
    \hline
        Reliance on dependencies between characteristics & C & \{None, Low, Medium, High \}\\
    \hline
        Ability to manage complex data & C & \{None, Low, Medium, High \}\\
    \hline
        Tolerance for biased data distributions & C & \{Low, Medium, High \}\\
    \hline
        Decision computational complexity & C & \{Low, Medium, High \}\\
    \hline
        Memory requirements & C & \{Low, Medium, High \}\\
    \hline
        Potential for parallelism/distribution & C & \{None, Partial, High \}\\
    \hline
        Support for federated learning & C & \{Low, Medium, High \}\\
    \hline
        Decision time boundedness (contrainte de temps) & C & \{Yes, No \} \\
    \hline
        Types of attributes & D & \{Categorical, Numerical, Textual \}\\
        \hline
    Evolutivity & D & \{Low, Medium, High \} \\
        \hline
    \end{tabular}
    \caption{Selection criteria, with weight and values ranges}
    \label{tab:criteres-selection-poids}
\end{table}

\subsection{Machine learning problem requirements}
\label{subsec:matching-problems-solutions:ML-problem-requirements}
Machine learning problem requirements can be divided into two subsets:
\begin{enumerate}
    \item Domain expert requirements: these represent requirements of the domain expert who has a domain problem to solve. Some of these requirements may be domain-wide. For example, in regulated industries such as financial services and insurance, auditability is important, which may, in turn, translate into \textit{Explainability} or \textit{Interpretability}, or a combination of the two. Other requirements may be specific to the application at hand.
    \item Properties of the data: these are requirements in the sense that they are part of the definition of the 'problem'. Whichever algorithm (family) we chose, has to be compatible with the characteristics of the data, including its type, volume, quality, and so forth. Some of these properties may be verifiable directly on the data, including \textit{Volume} and \textit{Type}. Others may require the input or judgement of the domain expert, such as \textit{Representativity}, \textit{Seasonality}, etc.
\end{enumerate}
The domain expert requirements are shown in Table \ref{tab:domain-expert-requirements}
\begin{table}
    \centering
    \begin{tabular}
    {| m{4cm} | m{4cm} | m{4cm} |}
        \hline
        \textbf{Requirement} & \textbf{Sub-requirements}  & \textbf{Value range}\\

    \hline  
    \hline
        Level of accuracy  &  Minimum requirement & a percentage\\
        & How much do you care & \{Not, Could, Should, Must \}\\
    \hline
        Explainability & Yes/No  & \{True, False \}\\
        & How much do you care & \{Not, Could, Should, Must \}\\
    \hline
        Interpretability & Yes/No & \{True, False \}\\
        & How much do you care & \{Not, Could, Should, Must \}\\
    \hline
        Adaptability with incremental change & Yes/No & \{True, False \}\\
        & How much do you care & \{Not, Could, Should, Must \}\\
    \hline
        Model training cost &  & \\
    \hline
        & Limit computations (CPU resources) & \{Low, Medium, High, Very high \}\\
        & How much do you care & \{Not, Could, Should, Must \}\\
    \hline
        & Limit data (acquisition costs) & \{Low, Medium, High, Very high \}\\
        & How much do you care & \{Not, Could, Should, Must \}\\
    \hline
        Decision speed & Max response time & a pair \{ metric\footnote{Could be Maximum(.), Average(.), Maximum(.) for 95\% of transactions, etc.}, duration\} \\
        & How much do you care & \{Not, Could, Should, Must \}\\
        \hline
    \end{tabular}
    \caption{Domain (expert) requirements}
    \label{tab:domain-expert-requirements}
\end{table}
Data properties are shown next. We showed for each property whether the value of the property can be determined automatically, given access to the data, or it needs to be supplied by the domain expert. For example, a property such as \texttt{volume} or \texttt{data type} can be assessed automatically, given access to the data. \texttt{Volume} can be measured in terms of number of records, or population size, or memory footprint. \texttt{Data type} can be assessed by inspecting values of data record fields/cells (numerical or else), or, by analyzing a data \textit{schema}, depending on the data representation technology. Our prototype ML problem solving platform, available at \url{https://isolvemymlproblem-c6d96d0c8560.herokuapp.com/login}, enables a domain expert to specify the database (URL and credentials) containing the training data for her/his ML problem. This enables to tool to access the database, and get its \texttt{data type} property value by querying its schema, and get its \texttt{volume} through a query. The property \texttt{missing values} may be evaluated in one of two ways: by querying the schema of the database to see if a given property is nullable, or by querying the attribute at hand checking for null values.

By contrast, a property such as \texttt{data labeling}, \texttt{seasonality} or \texttt{data rep\-re\-sentativity} is not intrinsic to the data itself, and cannot be checked automatically. The question of data labeling consists of identifying if a particular data field corresponds to the output that we want our trained model to predict, be it a diagnosis (i.e. a class label), a risk factor  (e.g. a regression), a review polarity, etc. Similarly for \texttt{seasonality}, unless we have historical data for several years--say--in which case we may be able to answer the question automatically, it is a property that needs to be supplied by the domain expert.
\begin{table}
    \centering
    \begin{tabular}
    {| m{4cm} | m{4cm} | m{4cm} |}
        \hline
        \textbf{Characteristic} & \textbf{Value range}  & \textbf{Comment}\\
    \hline  
        Data labeling  & \{Labeled, Unlabeled, To be labeled \} &  \\ 
    \hline
        Volume & volume metric & Can be checked automatically\\
    \hline
        Missing values & presence, absence, and sparseness & Can be checked automatically \\
    \hline
        Data type & \{Categorical, Numerical, Text \} & Can be checked automatically \\
    \hline
        Seasonality  & \{True, False \} & Ask the domain expert \\
    \hline
        Data representativity & \{Low, Medium, High\} & Within and between classes; ask the domain expert \\
    \hline
        \multirow{2}{4cm}{Homogeneity} & - Are the classes of comparable sizes & Ask the domain expert\\
        & - Do the attributes follow similar scales & Can be checked automatically \\
    \hline
        Data distribution & \{Normal, Unknown\} & Can be checked automatically \\
        \hline
    \end{tabular}
    \caption{Data requirements}
    \label{tab:data-requirements}
\end{table}

\subsection{Mapping requirements to algorithm (family) properties}
\label{subsec:matching-problems-solutions:mapping-requirements-selection-criteria}
The following table shows a mapping between \textit{problem requirements}, on one hand, and algorithm family properties (selection criteria), on the other. The problem requirements include the domain expert requirements shown in Table \ref{tab:domain-expert-requirements}, and the data properties shown in Table \ref{tab:data-requirements}.

As we can see, some domain requirements may map to several algorithm family properties. This is due to two factors: 
\begin{itemize}
    \item The inherent lack of \textit{precision} in the way that domain experts will express their needs. This is the case for \textit{adaptability with incremental change}, which translates into two separate qualities
    \item Genuine part-whole relationships between algorithm properties and domain requirements. This is the case for the \textit{computation} component of cost, which, for all practical purposes, can be mapped to different hardware components of a computer system: CPU, RAM, and compute/storage cluster.
\end{itemize}

\begin{table}
    \centering
    \begin{tabular}
    {| m{6cm} | m{6cm} |}
        \hline
        \textbf{Problem requirements} & \textbf{Algorithm (family) properties}\\

    \hline  
        Level of accuracy  & Accuracy (B) \\ 
    \hline  
        Interpretability  & Interpretability (A) \\ 
    \hline
    \multirow{2}{6cm}{Adaptability with incremental change} & Incrementality (C)\\
    & Evolutivity (D) \\
    \hline
        Cost & \\
        \hline
            \multirow{3}{6cm}{$-$ Computation} & Model training complexity (B) \\
        & Memory requirements (C) \\
        & Potential for parallelism/distribution (C) \\
        \hline
        $-$ Data & Memory requirements (C) \\
        \hline
        Decision speed & Decision complexity (C) \\
        \hline
        Data labeling & Training type (A) \\
        \hline
        Volume & Volume of data required for convergence (B) \\
        \hline
        Missing values & Tolerance for Missing full records or attribute values (B-C) \\
        \hline
        Data type (numerical vs. categorical) & Types of attributes (D) \\
        \hline
        Seasonality & Evolutivity (D) \\
        \hline
        Representativity & Tolerance to data imbalance (B) \\
        \hline
        Homogeneity & Tolerance to noise (C) \\
        \hline
        Data distribution & Tolerance for biased data distributions (C) \\
        \hline
    \end{tabular}
    \caption{Mapping problem requirements to algorithm family properties}
    \label{tab:problem-requirements-to-algorithm-properties}
\end{table}

\subsection{Desirable properties of the matching function}
\label{subsec:matching-problems-solutions:desirable-properties-matching-function}
To properly design a matching function between ML problems and algorithm families, we propose to:
\begin{enumerate}
    \item Elicit the desirable properties of our matching function, i.e. some sort of \textit{functional requirements} for the matching function; and then
    \item Propose a mathematical function that exhibits, a) those properties, \textit{first and foremost}, and b) any other mathematical properties that make the computation and interpretation of the results easier.
\end{enumerate}
We identify those functional requirements in this section; other convenient mathematical properties will be discussed in Section \ref{subsec:matching-problems-solutions:matching-function}.
\subsubsection{It depends}
\label{subsubsec:matching-problems-solutions:desirable-properties-matching-function:different-satisfaction-criteria}
Table \ref{tab:problem-requirements-to-algorithm-properties} shows which algorithm property is related to which problem requirement, but does not specify the nature of the relationship between the corresponding \textit{values}. We examine some examples to understand this relationship.

Let $Pb$ be an ML problem, $AF$ an algorithm family, and $Prop$ a property. A firsthand look at the matching problem would consider three generic functions, $Satisfies(AF, Prop)$, meaning that algorithm family $AF$ supports/exhibits property $Prop$, $Requires(Pb, Prop)$ meaning that problem $Pb$ requires property $Prop$, and $Solves(AF,Pb)$, meaning that algorithm family $AF$ can be used to solve the problem $Pb$. 

Given a set of properties $Prop_1$,..., $Prop_n$ such that $Requires(Pb, Prop_i)$, for i=1..n, we have:
\begin{equation}
\label{eq:boolean-interpretation-satisfaction}
    (\forall ~ AF) ~ Solves(AF,Pb) ~ \equiv  ~ \bigcap_1^n ~ Satisfies(AF,Prop_i )
\end{equation}
This characterization works for simple properties. For example, notwithstanding the 'how much you care' dimension (see Table \ref{tab:domain-expert-requirements}) a domain expert either requires \textit{interpretability}, or does not, and an algorithm family either supports it or does not.

Things get more complicated with \textit{Seasonality}--a data property--which maps to \textit{Evolutivity}, an algorithm family property. This would entail adding a fourth function $Map(Prop_{Pb,i}) \rightarrow Prop_{AF,i}$, such that:
\begin{multline}
     (\forall ~ AF)(\forall ~ Pb ~ s.t. \bigcap_1^n Requires(Pb, Prop_{Pb,i}))\\
 Solves(AF,Pb) ~ \equiv  ~ \bigcap_1^n ~ Satisfies(AF,Map(Prop_{Pb,i} ))   
\end{multline}
Regarding the \textit{Adaptability with incremental change} requirement, it actually maps to two algorithm family properties, \textit{Incrementality} and \textit{Evolutivity}. Things get even more complicated with (multi)valued properties, as opposed to binary ones, and 'fuzzy' requirements (e.g. \textit{low} computation cost) that map to several algorithm family \textit{numerical} properties.

From the above observations, we conclude that there is no single $Satisfies(AF$, $Prop_{Pb,i})$ function that applies across the various problem requirements (properties $Prop_{Pb,i}$). This means that we will have fifteen different satisfaction functions $Satis\-fies_{Prop_{Pb,i}}(Pb, AF)$, one for each requirement listed in (left column of) Table \ref{tab:problem-requirements-to-algorithm-properties}, that maps ($Pb$,$AF$) pairs to a numerical value that represents the extent to which $AF$ satisfies the requirements of $Pb$, with 0 meaning not at all and 1 meaning fully; these functions will be presented in Section \ref{subsec:matching-problems-solutions:matching-function}.

\subsubsection{It is not \textit{all} or \textit{nothing}}
\label{subsubsec:matching-problems-solutions:desirable-properties-matching-function:gradualness}
Our first attempt at formulating the matching function between ML problems and ML algorithm families (equation \ref{eq:boolean-interpretation-satisfaction}) framed the matching function as a boolean function that is a conjunction of the satisfaction functions for the individual properties ($Satisfies(AF,Prop_i )$). This means that if an algorithm family fails \textit{any} problem requirement, it is excluded. 

This would be too restrictive. For example, if we have categorical attributes in the data (\textit{Data type} requirement), this should not exclude algorithm families that expect numerical inputs, since we can always preprocess the data to map categorical values to numerical ones. This, plus the fact that the elementary satisfaction functions $Satisfies_{Prop_{Pb,i}}(Pb, AF)$ discussed previously return a value between 0 and 1, plead for a \textit{scoring} metaphor. One possible form is:
\begin{equation}
\label{eq:solves-weighted-score}
    Solves(AF,Pb) ~ \equiv  ~ \sum _1^n ~ W_i \times Satisfies_{Prop_{Pb,i}}(AF, Pb)
\end{equation}
where the weight $W_i$ represents the relative importance of the satisfaction of the problem requirement $Prop_{Pb,i}$. In the next two paragraphs, we explore two determinants of that weight.
\subsubsection{How much do you care?}
\label{subsubsec:matching-problems-solutions:desirable-properties-matching-function:how-much-you-care}
Domain expert requirements (see Table \ref{tab:domain-expert-requirements}) are qualified by 'how much [the domain expert] cares' about a particular requirement. For example, \textit{explainability} may be a must-have requirement for a ML-based mortgage underwriting decision, because of regulatory compliance, but no so much for an ML-powered ad-placement strategy for customer portals!  In table \ref{tab:domain-expert-requirements}, we used \textit{linguistic values} (\texttt{Not}, \texttt{Could}, \texttt{Should}, \texttt{Must}) for the 'how much do you care' dimension of domain expert requirements. These could be translated into numbers that reflect the relative reward, or penalty, for satisfying, or not, a given problem requirement.
\subsubsection{Does it matter?}
\label{subsubsec:matching-problems-solutions:desirable-properties-matching-function:discriminating-value}
Recall that the 'Weight' column in Table \ref{tab:criteres-selection-poids} is commensurate with the 'damage incurred', or amount of effort it takes to compensate, for failing the requirement. Thus, we can think of the weight $W_i$ in equation \ref{eq:solves-weighted-score} as:
\begin{multline}
\label{eq:weight-how-much-you-care-times-does-it-matter}
    W_i ~ \equiv ~ [ how ~ much ~ does ~ the ~ domain~ expert ~ care ~ about ~ Prop_{Pb,i} ]~ \times ~ \\
    [how ~ much~ the ~ corresponding ~ algorithm ~ property(ies) ~ matters]
\end{multline}
Notice that regarding \textit{data requirements} (Table \ref{tab:data-requirements}), the data characteristics, if known, are \textit{facts}. On a 0 to 1 'how much you care' scale, they correspond to a 1. With the exception of \textit{Data labeling}--a data characteristic that maps to algorithm family property \textit{Training type}, with weight $A$--the other data characteristics map to algorithm family properties mostly in the C and D category (Table \ref{tab:problem-requirements-to-algorithm-properties}).
\subsubsection{Normalization}
\label{subsubsec:matching-problems-solutions:desirable-properties-matching-function:normalization}
The function $Solves(AF,Pb)$ presented earlier represents the extent to which algorithm family $AF$ is a good fit for the problem $Pb$. Because different domains, and domain experts, may have different sets of requirements, we should normalize the function. Thus, we should divide the value of $Solves(AF,Pb)$ in equation \ref{eq:solves-weighted-score} by the sum of weights. Accordingly, we now define $Solves(AF,Pb)$ as follows:
\begin{equation}
\label{eq:solves-normalized-score}
    Solves(AF,Pb) ~ \equiv  ~ { \sum _1^n ~  W_i \times Satisfies_{Prop_{Pb,i}}(AF, Pb)  \over{\sum_1^n ~ W_i}}
\end{equation}
\subsection{The matching function}
\label{subsec:matching-problems-solutions:matching-function}
In this section, we present the different $Satisfies_{Prop_{Pb,i}}(AF, Pb)$ functions; the overall $Solves(AF,Pb)$ being given by the equation \ref{eq:solves-normalized-score}.
\subsubsection{Satisfying the accuracy level requirement}
\label{subsubsec:matching-problems-solutions:matching-function:accuracy}
For accuracy, the higher the better. But because higher accuracy levels come with a cost, the \textit{Level of accuracy} requirement is meant to indicate the minimum level of accuracy that is acceptable for the problem at hand. Typically, the minimum acceptable level would be lower for an ad placement application for the landing page of a customer portal, then for a cancer diagnosis application. However, any algorithm family beyond the minimum requirement would do. Thus:
\begin{multline}
\label{eq:satisfies-level-accuracy-binary}
(\forall~problem ~ Pb, ~ \forall ~ algorithm ~ family ~ AF)\\ 
Satisfies_{accuracy} (AF,Pb) ~ \equiv ~ Accuracy(AF) ~ >=~ LevelAccuracy(Pb)
\end{multline}
This function is binary. A more gradual satisfaction function follows:
\begin{multline}
\label{eq:satisfies-level-accuracy-gradual}
Satisfies_{accuracy} (AF,Pb) ~ \equiv ~ Minimum \left( 1, {Accuracy(AF) \over {LevelAccuracy(Pb)}} \right)
\end{multline}
\subsubsection{Satisfying the interpretability requirement}
\label{subsubsec:matching-problems-solutions:matching-function:interpretability}
This is a 'binary' requirement: it is either required by the domain (expert) or not. If it is not required, the function $Satisfies_{Interpretability}(.,.)$ will not be included in the computation of the $Solves(AF, Pb)$ function. 
\begin{multline}
\label{eq:satisfies-interpretability}
(\forall~problem ~ Pb ~ s.t. ~ Interpretability(Pb), ~ \forall ~ algorithm ~ family ~ AF)\\ 
Satisfies_{interpretability} (AF,Pb) ~ \equiv ~
\begin{cases}
    1, & Interpretable(AF) \\
    0, & otherwise
\end{cases}
\end{multline}
\subsubsection{Satisfying the adaptability requirement}
\label{subsubsec:matching-problems-solutions:matching-function:adaptability}
The adaptability requirement maps to two algorithm family properties, with different weights: \textit{Incrementality} (C), and \textit{Evolutivity} (D). If we think of it as a weighted average of the two properties:
\begin{multline}
\label{eq:satisfies-adaptability}
(\forall~problem ~ Pb ~ s.t. ~ AdaptableIncrementalChange(Pb), ~ \forall ~ AF)\\ 
Satisfies_{Adaptability} (AF,Pb) ~ \equiv ~ \\
W_C \times Incremental(AF) ~+~ W_D \times Evolutive(AF) \over {W_C~+~W_D}
\end{multline}
where $W_C$ and $W_D$ are the numerical values associated with the weights $C$ and $D$, respectively, given to algorithm properties; see Sections \ref{subsec:matching-problems-solutions:selection-criteria-algorithm}, and \ref{subsec:matching-problems-solutions:desirable-properties-matching-function}.
\subsubsection{Satisfying the cost requirement}
\label{subsubsec:matching-problems-solutions:matching-function:cost}
A \textit{low} cost is always desirable. The \textit{Model training cost} requirement (Table \ref{tab:domain-expert-requirements}) was divided into CPU and memory costs. It represents an upper limit of the cost that a domain (expert) will tolerate. The cost limitations may be \textit{soft}, financial constraints, or physical, \textit{hard} constraints, based on the capacities of the actual or potential devices that will do the training. 

Cost characteristics are difficult to specify precisely. On the requirements side, few domain experts are able to specify the hardware limitations of their platforms. With algorithm families, we can express computational complexity in 'big O' notation, which indicates growth trends, not actual values. Further, they often represent theoretical limits. Accordingly, we choose to specify both cost requirements and algorithm family complexity using \textit{linguistic values} (\texttt{Low}, \texttt{Medium}, \texttt{High}, \texttt{Very High}).

First, at a high level, the $Satisfies_{cost}(AF,Pb)$ function can be written as a combination of two subfunctions,  $Satisfies_{cost ~ CPU}(AF,Pb)$, and $Satis\-fies_{cost ~ memory}(AF,Pb)$, where \textit{CPU} stands for computation-related costs:
\begin{multline}
\label{eq:satisfies-cost}
(\forall ~ Pb, ~ \forall ~ AF) ~~ Satisfies_{cost} (AF,Pb) ~ \equiv ~\\
{ Satisfies_{cost~ CPU}(AF,Pb) ~+~ Satisfies_{cost~ memory}(AF,Pb)  \over {2} }
\end{multline}
From Table \ref{tab:problem-requirements-to-algorithm-properties}, we see that computational costs are related to \textit{Model training complexity} (B), \textit{Memory requirements} (C), and \textit{Potential for parallelism} (C). Each property is a function that takes an algorithm family as an argument and returns a linguistic value (\texttt{Low}, \texttt{Medium}, \texttt{High}, and \texttt{Very High}).

Assuming the order \texttt{Low} $\leq$ \texttt{Medium} $\leq$ \texttt{High} $\leq$ \texttt{Very High}, we define a function $\leq_{fuzzy}(v_1,v_2)$, that returns the extent to which $v_1$ is smaller than $v_2$,  as follows:
\begin{multline}
\label{eq:less-fuzzy}
\leq_{fuzzy}(v_1,v_2) ~ \equiv ~  Min\left(1, 1 - {rank(v_1) ~-~rank(v_2)  \over{3}}\right)
\end{multline}
Thus, $\leq_{fuzzy}(Low,Low)$ = $\leq_{fuzzy}(Low,Medium)$ = $\leq_{fuzzy}(Low$, $High)$ = 1. However, $\leq_{fuzzy}(Medium,Low)$ = $\leq_{fuzzy}(High,Medium)$ = $\leq_{fuzzy}(Very~High, High)$ = $2/3$, $\leq_{fuzzy}(High$, $Low)$ = $\leq_{fuzzy}(Very ~ High$, $Medium)$ = $1/3$, and $\leq_{fuzzy}(Very ~ High,Low)$ = 0. We can now compute $Satisfies_{cost~ CPU}(AF,Pb)$ as follows:
\begin{multline}
\label{eq:satisfies-cost-CPU}
\forall ~ Pb, ~ \forall ~ AF) ~~ Satisfies_{cost~CPU} (AF,Pb) ~ \equiv ~ {1 \over {W_B ~+~ 2 \times W_C}} ~ \times \\
[ W_B \times \leq_{fuzzy}(ModelTrainingComplexity(AF),ComputationCost(Pb)) ~+~ \\
W_C \times \leq_{fuzzy}(MemoryRequirements(AF),ComputationCost(Pb)) ~+~ \\
W_C \times \leq_{fuzzy}(Complement(PotentialParallelisation(AF)), \\
ComputationCost(Pb)) ]
\end{multline}
where $Complement(V)$, for a given linguistic value, returns the value with rank $5 ~- rank(V)$. Thus, $Complement(Low)$ =  \texttt{Very High} (and vice-versa), and $Complement(High)$ =  \texttt{Medium} (and vice-versa). This to account for the fact that high potential for parallelisation is a good thing, that is compatible with a desire for low computation cost.

The function $Satisfies_{cost~ memory}(AF,Pb)$ is defined as follows:
\begin{multline}
\label{eq:satisfies-cost-memory}
(\forall ~ Pb, ~ \forall ~ AF) ~~ Satisfies_{cost~memory} (AF,Pb) ~ \equiv ~ \\
\leq_{fuzzy}(MemoryRequirements(AF),MemoryCost(Pb)) 
\end{multline}
\subsubsection{Satisfying the remaining requirements}
\label{subsubsec:matching-problems-solutions:matching-function:remaining-requirements}
The previous paragraphs showed the range of relationships that can exist between problem requirements (or data characteristics) on one hand, and algorithm family properties, on the other. 

For the purposes of this paper, space limitations do not allow us to go over the remaining nine properties, but they each correspond to one of the four patterns seen previously. For example, the discussion pertaining to \textit{Interpretability} showed the pattern for \textit{binary} requirements, and properties. That pattern, embodied in equation \ref{eq:satisfies-interpretability}, can be used for the \textit{Seasonality} data characteristic, which maps to the \textit{Evolutivity} property for algorithm families. Thus, the $Satisfies_{Seasonality} (AF,Pb)$ function can be defined as follows:
\begin{multline}
\label{eq:satisfies-seasonality}
(\forall~problem ~ Pb ~ s.t. ~ Seasonality(Pb), ~ \forall ~ algorithm ~ family ~ AF)\\ 
Satisfies_{Seasonality} (AF,Pb) ~ \equiv ~
\begin{cases}
    1, & Evolutive(AF) \\
    0, & otherwise
\end{cases}
\end{multline}
If we map \texttt{true} to 1 and \texttt{false} to 0, we get the simpler formulation:
\begin{multline}
\label{eq:satisfies-seasonality}
(\forall ~ Pb ~ s.t. ~ Seasonality(Pb), ~ \forall ~ AF)\\ 
Satisfies_{Seasonality} (AF,Pb) ~ \equiv ~ Evolutive(AF)
\end{multline}
The full matching function ($Solves(AF,Pb)$) is provided in \cite{Saleh2024}.
\section{Related Work}
\label{sec:related-work}
Several research works have sought to demystify the use of ML techniques to solve common problems. They can be divided into four categories. First, there is work focused on the presentation of ML algorithms or algorithm families, along with advantages and disadvantages. The corresponding publications tend to be aimed at an audience of specialists to help them select (manually) the algorithm family for a particular class of problems, and configure its parameters (e.g. \cite{57das2017survey, fatima2017}). Our work builds on this line of work, including as a source of ML algorithm families and selection criteria (Section \ref{subsec:matching-problems-solutions:selection-criteria-algorithm}).

A second category of work aims at providing practitioners with simplified guides to help them select the right algorithms. These guides, known as \textit{cheat sheets}, are typically given as \textit{decision trees} that help the user navigate families of ML algorithms based on a sequence of simple questions.  Examples include Microsoft Azure \cite{Microsoft}, Scikit Learn \cite{Scikit}, and SAS \cite{SAS2020}.  While helpful, these cheat sheets are limited in many ways. First, they focus on a single step of the solution process: the algorithm family to use for model building/training; there is a lot more to ML problem solving that algorithm family selection. Second, they do not recognize algorithm selection as an optimization problem that involves trade-off (e.g. accuracy versus explainability). 

A third category of relevant work, called \textit{Auto ML}, deals with \textit{fully automating} the design of ML solutions for domain problems, \textit{using machine learning}. Examples include AutoWeka \cite{thornton2013}, AutoSklearn \cite{feurer2015}, MLPlan \cite{mohr2018}, AutoGluon \cite{AutogluonRN29}, AutoStacker \cite{chen2018}, and others. In practice, the method is not scalable due to the size of the search space. Further, it focuses solely on the model training part of the ML-based solution. Finally, the entire process is a black box. 

Our work fits in a fourth, \textit{long} line of research, which aims at providing \textit{semi}-automated \textit{assistance} in the selection and configuration of algorithms, including \cite{AkaikeRN51}, \cite{KotsiantisRN54}, \cite{RayRN81}, and others. These approaches do \textit{not} attempt to automate the full problem solving process. However, they codify problem solving expertise in automated procedures or optimization algorithms, i.e. \textit{procedural knowledge}, which lacks transparency and incrementality. By contrast, our approach encodes problem solving expertise in \textit{transparent data} (ML solution artefacts and their properties), with simple and transparent processes (querying, scoring, etc.).
\section{Validation}
\label{sec:validation}
\subsection{Validation strategy}
\label{subsec:validation:validation-strategy}
We follow a Design Science Research Methodology (DSRM) \cite{Hevner2004}. Our research aims at developing tools for domain specialists to help them develop "good enough" solutions to common domain problems using machine learning (ML) techniques. We rely on a combination of: 1) an appropriate \textit{representation} of domain problems, ML solutions patterns, and ML technical artefacts, 2) an inventorying and codification of ML solution patterns and ML artefacts, and 3) a number of \textit{simple} tools to help with the design of individual problem solutions through search, specialization, and composition of various solution artefacts.

Ultimately, the value of our approach resides in the extent to which the platform that we are developing does help domain experts who are not machine learning experts, to produce "good enough" solutions for common domain problems using ML techniques. However, at this point in the research, we are not yet at a stage where we can evaluate our problem solving workbench (see \url{https://isolvemymlproblem-c6d96d0c8560.herokuapp.com/login}).

In DSRM terminology, the workbench is but one \textit{artefact}, referred to as \textit{instantiation}, which is the end product aimed at solving the original problem \cite{Hevner2004}: providing ML solving assistance to non ML experts. Earlier (by)products of DSRM are:
\begin{itemize}
    \item \textit{Constructs}, which "provide the language in which problems and solutions are defined and
communicated" \cite{Hevner2004}. In our case, the \textit{constructs} consist of the ML project metamodel presented in Subsection \ref{subsec:model-ML-problems-solutions:ML-project-metamodel}, and the representation of algorithms, algorithm families, and processing chains (Subsection \ref{subsec:model-ML-problems-solutions:representing-algorithms-processing-pipelines}).
\item \textit{Models}, which "use constructs to represent ... the design problem and its solution space" \cite{Hevner2004}. In our case, \textit{models} are embodied in, a) the \textit{selection criteria} presented in Subsection \ref{subsec:matching-problems-solutions:selection-criteria-algorithm}, b) the algorithm families that we choose to populate the solution artefact database (see Figures \ref{fig:ML-problem-solving-workflow} and \ref{fig:concept-of-operations}), and c) \textit{values} for those selection criteria, for those algorithm families.
\item \textit{Methods}, which "provide guidance on how to solve problems, that is, how to search the
solution space" and that can be "formal, mathematical algorithms that explicitly define the
search process" \cite{Hevner2004}. Our matching function (Subsections \ref{subsec:matching-problems-solutions:desirable-properties-matching-function} and \ref{subsec:matching-problems-solutions:matching-function}) is one such \textit{crucial} method, but not the only one.
\end{itemize}
Per Hevner et al., the \textit{instantiations} put constructs, models and methods in a working system intended to solve the original problem \cite{Hevner2004}.
The remainder of this section will outline our strategy for validating the different artefacts. \textit{Constructs} will be validated within the context of the validation of \textit{models}.
\subsection{Validating models}
\label{subsec:validation:validating-models}
As mentioned above, we must validate three aspects: a) selection criteria, b) our selection of algorithm families to populate the tool's solution artefact database, and c) the \textit{values} used for the selection criteria for the selected program families.

Regarding the selection criteria, Section \ref{subsec:matching-problems-solutions:selection-criteria-algorithm} explained the process used to identify the list of selection criteria used to describe program families. As explained in Section \ref{subsec:matching-problems-solutions:selection-criteria-algorithm} the starting point was a thorough literature review\footnote{Which is part of a systematic review (in progress) on ML problem solving aids; see the related works section (\ref{sec:related-work}) for a glimpse.}. There then followed a number of steps to eliminate redundancies, dependent properties, and non-salient properties, where the authors used their judgement and ML knowledge. Those judgements will need external validation.

Regarding the identification of program families to include in the database of the workbench, there are hundreds, if not thousands, of published algorithm families and algorithm variants; for the purposes of our workbench, we chose to not go beyond two dozen basic program families so as to not overwhelm the intended users, who are not ML specialists \cite{Saleh2024,SalehICIEA2024}. There is a wide consensus in the literature about such high-level algorithm families might be: they are typically listed in the major \textit{cheat sheets} (e.g. \cite{Microsoft}, \cite{Scikit}, SAS \cite{SAS2020}). However, differences may exist in the way they are hierarchically organized--e.g., by learning style, \textit{first}, vs shape of the ML model.

Regarding the \textit{values} of the selection criteria for the chosen algorithm families, we are relying on two complementary sources: 1) the scientific literature, and 2) \textit{crowd sourcing} \cite{SalehICIEA2024}. Several of the selection criteria are factual/provable (e.g. \textit{decision complexity}) or non controversial (e.g. learning type). Others may be more observational/empirical where, short of a mathematical proof, different authors may arrive at different assessments. For example, one author might characterize a particular algorithm family as having a \texttt{High} \textit{tolerance to noise} while another, using the same algorithm family on a different problem, might assign the value \texttt{Medium}. For this reason, we published a platform called \textit{icontributetoml}, that enables registered users to enter their own values for the selection criteria of algorithm families, and to justify them with text and bibliographic references \cite{Saleh2024,SalehICIEA2024}.

Space limitations do not allow us to go into the details of \textit{icontributetoml}; for more details, see \cite{Saleh2024,SalehICIEA2024}. Suffice it to say that the \textit{icontributetoml} platform aims to \textit{help} us capture the consensus of the ML community, guaranteeing \textit{some} level of \textit{model validity} 'by design'. However, because we cannot select the users, we are working on complementing this 'crowd sourcing' effort with formal controlled experiments where we select the participants, and a sample of the selection criteria, to validate for the algorithm families represented in the tool.
\subsection{Validating methods}
\label{subsec:validation:validating-methods}
A main design driver of our workbench is that it is \textit{not} intelligent: the design aids that the workbench provides rely on queries and simple computations, such as the ones included in the matching function (Subsection \ref{subsec:matching-problems-solutions:matching-function}). The "intelligence" is in the "data": having the appropriate \textit{schema} (the metamodels described in Section \ref{sec:model-ML-problems-solutions}), and the appropriate \textit{values} for the selection criteria.

At this point in the research, we have fully designed and implemented the matching function between ML problem requirements and algorithm family properties (Section \ref{sec:matching-problems-solutions}). We have designed an experimental protocol to validate the matching function, and started executing some of its steps (details in \cite{Saleh2024}):
\begin{enumerate}
    \item Build a dataset of ML problems from real case studies published in the scientific and professional literature. The ML problems need to cover a large enough spectrum of requirements (domain and data requirements), in terms of presence or absence (e.g. need for explainability), values ranges, and priorities ('how much you care', see Subsection \ref{subsec:matching-problems-solutions:desirable-properties-matching-function}).
    \item Recruit experimental subjects with proficiency in the \textit{main} machine learning algorithm families. The proficiency level is assessed with a written exam, at the level of a final exam for a graduate course in machine learning.
    \item Provide the experimental subjects with a set of ML problems and have them specify, for each problem, a maximum of five algorithm families, ranked by order of fitness for the requirements at hand.
    \item Use the matching function described in Section \ref{sec:matching-problems-solutions} to score and rank the algorithm families against each of the ML problem requirements.
    \item Using ranking correlation metrics, to: a) assess inter-subject agreement, and when possible, b) assess the agreement between the human experimental subjects, and the matching function.
\end{enumerate}
We are currently doing (1) and (2). We are also working with a statistician to finalize various experimental parameters and protocols, depending on the size of the data set and the number of successfully recruited participants (see \cite{Saleh2024}).
\subsection{Validating the \textit{isolvemymlproblem} workbench}
\label{subsec:validation:validating-isolvemymlproblem}
The matching function between ML problem requirements and algorithm families is but \textit{one} method, used in the \textit{first} step of the problem solving workflow described in Section \ref{sec:towards-ML-based-solution-workbench}. Other \textit{methods} are at various stages of design, implementation, and testing \cite{Saleh2024}. 
For the purposes of this paper, suffice it to say that:
\begin{itemize}
\item An evaluation of \textit{isolvemymlproblem} must assess \textit{functionality} and \textit{usability}.
\item \textit{Functionality} is concerned with the extent to which the ML problem solution is a 'good' one--'good' to be defined.
\item \textit{Usability} is concerned with whether the tool uses appropriate \textit{metaphors}, for the intended user community (e.g., a doctor), to elicit problem requirements.
\end{itemize}
Regard functionality, the validation protocol will be significantly more complex than that for the matching function (Section \ref{subsec:validation:validating-methods}), for at least two reasons: 
\begin{enumerate}
    \item The \textit{isolvemymlproblem} platform provides design aids, and different users may use different aids. If we let them, the results will be difficult to compare; if we do not, the experiment will not be realistic.
    \item An ML solution is a \textit{processing chain} (Section \ref{sec:towards-ML-based-solution-workbench}, \ref{subsec:model-ML-problems-solutions:concept-of-operations}, \ref{subsec:model-ML-problems-solutions:representing-algorithms-processing-pipelines}), involving a more or less complex assembly of components, the model training one being just one of many. This makes assessing or comparing processing chains complex.
\end{enumerate}
\section{Discussion}
\label{sec:discussion}
Our research aims at developing tools for subject matter experts to help them develop good enough solutions to common domain problems using machine learning (ML) techniques. We rely on a combination of: 1) an appropriate \textit{representation} of domain problems, ML solutions patterns, and ML technical artefacts, 2) an inventorying and codification of ML solution patterns and ML artefacts, and 3) a number of \textit{simple} tools to help with the design of individual problem solutions through search, specialization, and composition of various solution artefacts.

We adopted the Design Science Research Methodology (DSRM) \cite{Hevner2004}, and we are proceeding incrementally for both the development and the validation of the new artefacts. Our strategy for codifying solution artefacts uses a combination of a literature review, and a crowd-sourcing process and tool to build domain artefacts at scale. Much work remains to be done to make it a success.
\bibliographystyle{splncs04}
\bibliography{representation-and-matching}
\end{document}